%% file: AAA_V2.tex
\documentclass[10pt, conference]{IEEEtran}
\IEEEoverridecommandlockouts
\usepackage{epsfig}
\usepackage{graphicx}
\usepackage{amsmath}
\usepackage{amssymb}
\usepackage{float}
\usepackage{tabularx}
\usepackage{bbm}
\usepackage{breqn}
\usepackage{booktabs}
\usepackage{adjustbox}

\usepackage[pagebackref=true,breaklinks=true,colorlinks,bookmarks=false]{hyperref}

\ifCLASSINFOpdf
\else
\fi

\usepackage{graphicx}  

\usepackage{booktabs} 

\usepackage{adjustbox}

\usepackage{tikz}
\usetikzlibrary{shapes, arrows.meta, fit}

\tikzstyle{decision} = [diamond, draw, fill=Maroon!50, 
text width=4em, text badly centered, node distance=3cm, inner sep=0pt]
\tikzstyle{block} = [rectangle, draw, fill=Maroon!50, 
text width=5em, text centered, rounded corners, minimum height=3em]
\tikzstyle{block2} = [rectangle, draw, fill=Goldenrod!20, 
text width=2em, text centered, rounded corners, minimum height=2em]
\tikzstyle{block3} = [rectangle, draw, fill=Goldenrod!20, 
text width=4em, text centered, rounded corners, minimum height=2em]
\tikzstyle{line} = [draw, -latex']
\tikzstyle{cloud} = [draw, ellipse,fill=BurntOrange!50, node distance=3cm,
minimum height=1em]
\tikzstyle{black} = [draw, regular polygon, regular polygon sides=10, minimum height=1em, minimum width=2em]
\tikzstyle{white} = [draw, circle, minimum height=1em]

\definecolor{Goldenrod}{RGB}{249, 214, 22}
\definecolor{BurntOrange}{RGB}{199, 98, 43} 	   
\definecolor{Maroon}{RGB}{152, 30, 50}

\begin{document}

\title{Preliminary Studies on a Large Face Database}


\author{
\IEEEauthorblockN{Benjamin Yip}
\IEEEauthorblockA{Mathematical Decision Sciences\\
UNC Chapel Hill\\
NC, United States\\
benyip@live.unc.edu}
\and
\IEEEauthorblockN{Garrett Bingham\footnotemark{*}}
\IEEEauthorblockA{Computer Science 
\\
Yale University\\
CT, United States\\
garrett.bingham@yale.edu}
\thanks{*equal authorship}
\and
\IEEEauthorblockN{Katherine Kempfert\footnotemark{*}}
\IEEEauthorblockA{Statistics and Mathematics\\
University of Florida\\
FL, United States\\
kkempfert2@ufl.edu}
\and
\IEEEauthorblockN{Jonathan Fabish}
\IEEEauthorblockA{Applied Mathematics\\
N.C. A\&T State University\\
NC, United States\\
jlfabish@gmail.com}
\and
\IEEEauthorblockN{Troy Kling, Cuixian Chen, Yishi Wang}
\IEEEauthorblockA{Mathematics and Statistics\\
UNC Wilmington\\
NC, United States\\
\{tpk7509, chenc, wangy\}@uncw.edu}
}

\maketitle

\begin{abstract}
We perform preliminary studies on a large longitudinal face database MORPH-II, which is a benchmark dataset in the field of computer vision and pattern recognition. First, we summarize the inconsistencies in the dataset and introduce the steps and strategy taken for cleaning. The potential implications of these inconsistencies on prior research are introduced. Next, we propose a new automatic subsetting scheme for evaluation protocol. It is intended to overcome the unbalanced racial and gender distributions of MORPH-II, while ensuring independence between training and testing sets. 
Finally, we contribute a novel global framework for age estimation that utilizes posterior probabilities from the race classification step to compute a race-composite age estimate. Preliminary experimental results on MORPH-II are presented. 
\end{abstract}

\IEEEpeerreviewmaketitle

%

%



%

%
%
%
\section{Introduction}
MORPH is one of the largest publicly available longitudinal face databases \cite{ricanek2006morph}. 
Multiple versions of MORPH have been released. In this paper, the 2008 non-commercial release will be considered and referred to as MORPH-II. The MORPH-II dataset is a collection of 55,134 mugshots taken between 2003 and late 2007.
It includes many images of individuals who were arrested multiple times over the five year span. On average, there are approximately 4 images per subject. Additionally, there are images with pose, lighting, and/or expression variations, along with occlusion. Because of its longitudinal span, large number of subjects, variation in images, and inclusion of relevant metadata, MORPH-II has become one of the benchmark datasets in the field of computer vision and pattern recognition. It has been used for a variety of applications in face recognition and demographic analysis. In particular, the MORPH-II dataset is widely utilized in research on gender \cite{han2015demographic} and race classification \cite{guo2010study}, age estimation \cite{antipov2017effective, jana2017automatic, niu2016ordinal, liu2015age, guo2010human}, and age synthesis \cite{fu2010age}. However, there are some challenges with MORPH-II that we address here.

The first issue with MORPH-II is inconsistency (identified from our exploratory data analytics) in records of subjects' age, gender, and race.
To our best knowledge, no previous work with the MORPH-II dataset has acknowledged these inconsistencies, which can be critical in demographic analysis such as gender and race classification, as well as age estimation and age synthesis. Accordingly, the first goal of this paper is to investigate the inconsistencies in MORPH-II and to detail our cleaning methodology. We believe this preliminary study can provide a general guide to the data validation and data cleaning process for future studies on MORPH-II. 

Another challenge with MORPH-II is creating balanced subsets for cross-validation (CV), a popular approach for estimation of prediction errors. One complication with CV on MORPH-II is information leaking, which could potentially inflate an algorithm's performance by allowing a subject's images in both the training and testing sets. Additionally, MORPH-II is highly imbalanced towards black male subjects. To address such challenges, Guo and Mu \cite{ guo2010human, guo2010study, guo2011simultaneous} proposed an evaluation protocol that has been adopted by many studies on MORPH-II. An issue with their subsetting scheme is that many individuals arrested multiple times have images included in both training and testing. Moreover, it can be tedious and time-consuming to manually split the dataset into subsets satisfying the necessary criteria. Therefore, the second goal of this paper is to investigate how to {\it automatically} split MORPH-II into non-overlapping folds following the evaluation protocol proposed by Guo and Mu \cite{ guo2010human, guo2010study, guo2011simultaneous}.

After attempting to address these issues with MORPH-II, we consider the task of age estimation, which has applications ranging from Electronic Customer Relationship Management (ECRM) to surveillance monitoring and biometrics \cite{survey}. Age estimation has been widely researched in the field of computer vision and pattern recognition. Previous research \cite{guomu} has shown that age estimation is highly sensitive to race and gender categories, informing much of the framework proposed in this paper. While models with race and gender dependent age predictions have been explored, to our best knowledge, none have accounted for individuals of mixed race. In the real world, where diversity within society is increasing, failure to account for multiracial individuals could create unrealistic models that undermine practical applications. To address this discrepancy, we propose a model utilizing posterior probabilities from the race classification step for age estimation.

The paper is organized as follows: Section II presents our preliminary study on the inconsistencies and cleaning of MORPH-II. Our proposed automatic subsetting scheme is presented in Section III.
A global framework of race-composite age estimation with posterior probabilities from the race classification step is introduced in Section IV. Conclusions are drawn in the final section of this paper.

\section{Inconsistencies and Cleaning in MORPH-II}
We account the inconsistencies in the MORPH-II dataset and introduce the steps and strategy taken to clean it.

\subsection{Inconsistencies }
We found that the true number of unique individuals in the MORPH-II dataset is 13,617 (by using the 6-digit subject identifier {\it id\_num}). However, there are 11,459 unique males and 2,159 unique females in the dataset, which makes the total number of distinct subjects by gender 13,618. This suggests there may be an individual listed as both male and female. Repeating the same procedure for race produced similar results; as shown in Table \ref{Table-unique-race}, the total number of distinct individuals (13,658) does not agree with the true number of unique individuals (13,617). To investigate age, we compared the recorded dates of birth for each subject and found 1,779 subjects with inconsistent birthdates. These inconsistencies among gender, race, and birthdate are summarized in Table \ref{tab_inconsistent}.
\begin{table}[ht]
	\centering
    \caption{Number of Distinct Individuals by Race and Gender}
    \label{Table-unique-race}
	\begin{tabular}{lcccccc}
        \toprule
                & \textbf{B}lack  & \textbf{W}hite  & \textbf{A}sian & \textbf{H}ispanic & \textbf{O}ther & \textbf{Total}  \\ \midrule
\textbf{Male}   & 8,838  & 2,070  & 49    & 517      & 15    & 11,489 \\
\textbf{Female} & 1,494  & 634   & 6     & 30       & 5     & 2,169  \\
\textbf{Total}  & 10,332 & 2,704  & 55    & 547      & 20    & \textbf{13,658}\\ \bottomrule
	\end{tabular}

\bigskip
      \caption{MORPH-II Inconsistencies by Attribute}
      \label{tab_inconsistent}
      \begin{tabular}{ll}
      \toprule
      Attribute & Number of Subjects \\ \midrule
      Gender    & 1                \\
      Race      & 33               \\
      Birthdate & 1,779             \\ \bottomrule
      \end{tabular}
    \end{table}


These inconsistencies arise from subjects who were arrested multiple times over the five year span of MORPH-II. Indeed, all but 457 subjects were arrested multiple times, with an extreme case of one individual arrested 53 times. 
Some of these subjects have more than one gender, race, and/or birthdate reported across their database entries. Most data gathered for mugshots are self-reported with technological verification of the information, and evidently some errors have occurred. This may cause critical problems if MORPH-II images are used to build facial demographic systems, such as age estimation or race classification. Note that for the 457 subjects with only one entry in the dataset, there is no way to check whether the reported information is correct or not.

\subsection{Cleaning Process}
In this section, we detail the methods used to resolve the inconsistencies in MORPH-II.

\subsubsection{Cleaning for Gender Inconsistency}
There is only one subject with inconsistent gender in the database, listed as both male and female. However, this individual was reported as male during only one arrest, while a prior arrest and three subsequent arrests reported female (Figure \ref{gender_inconsistent}). As a result, we assumed this inconsistency was a clerical error and corrected the subject's gender to be female in all images.

\input{Inconsistent_image.tex}

\subsubsection{Cleaning for Race Inconsistencies}
There are 33 subjects with 132 images in MORPH-II with inconsistent race.We note that subjects who identify as mixed race are included in the Other race category in MORPH-II.
In order to best determine race in the MORPH-II dataset, human perception was utilized. To reduce personal bias, a group of researchers were trained by the following steps. First, literature on race classification was carefully selected and reviewed, including \cite{guo2010study}, \cite{guo2010human}, and \cite{fu2014learning}. The literature outlines the significance of eyes and nose and the insignificance of features such as skin tone. Researchers were also made aware of possible bias from the other-race-effect: the tendency to more easily recognize faces of one's own race. The most popular and effective methods of perceiving race were summarized to create race perception guidelines. Next, researchers were trained on human race perception with correctly labeled images in MORPH-II. The images of Asians and Hispanics were a focus, as these represented the majority of the misclassifications.

After reviewing the literature and undergoing training in race perception, the researchers attempted to identify the race of the subjects in question. To start, any individual with a clear majority of images belonging to one race was identified as this majority race. For example, the first subject in Figure \ref{race_inconsistent} has 24 images classified as White, while 1 image is classified as Black. Thus, this subject was identified as White and image (1b) was relabeled as White. In cases without a clear majority, human race perception was applied. Finally, in the case that not enough information was available or the race of an individual was unclear or mixed race, the subject was identified as the Other race category. For example, the second subject in Figure \ref{race_inconsistent} was identified as Other because she did not clearly exhibit only one race, according to the researchers trained in race perception.


\subsubsection{Cleaning for Birthdate Inconsistencies}
Of the 1,779 subjects in MORPH-II with inconsistent birthdates, 1,524 were identified and resolved with a simple majority, much like with person 1 in Figure \ref{Worst Inconsistent Birthdates}. However, the remaining 255 subjects pose additional problems. For some of them, there is no majority, or their birthdates differ by several years. This presents difficulties in choosing one birthdate over another.

\begin{figure}[h]
\centering
\caption{Cumulative proportion of subjects with inconsistent birthdates by number of days.} \bigskip
\label{age_inconsistent}
\includegraphics[width = 0.48\textwidth,keepaspectratio]{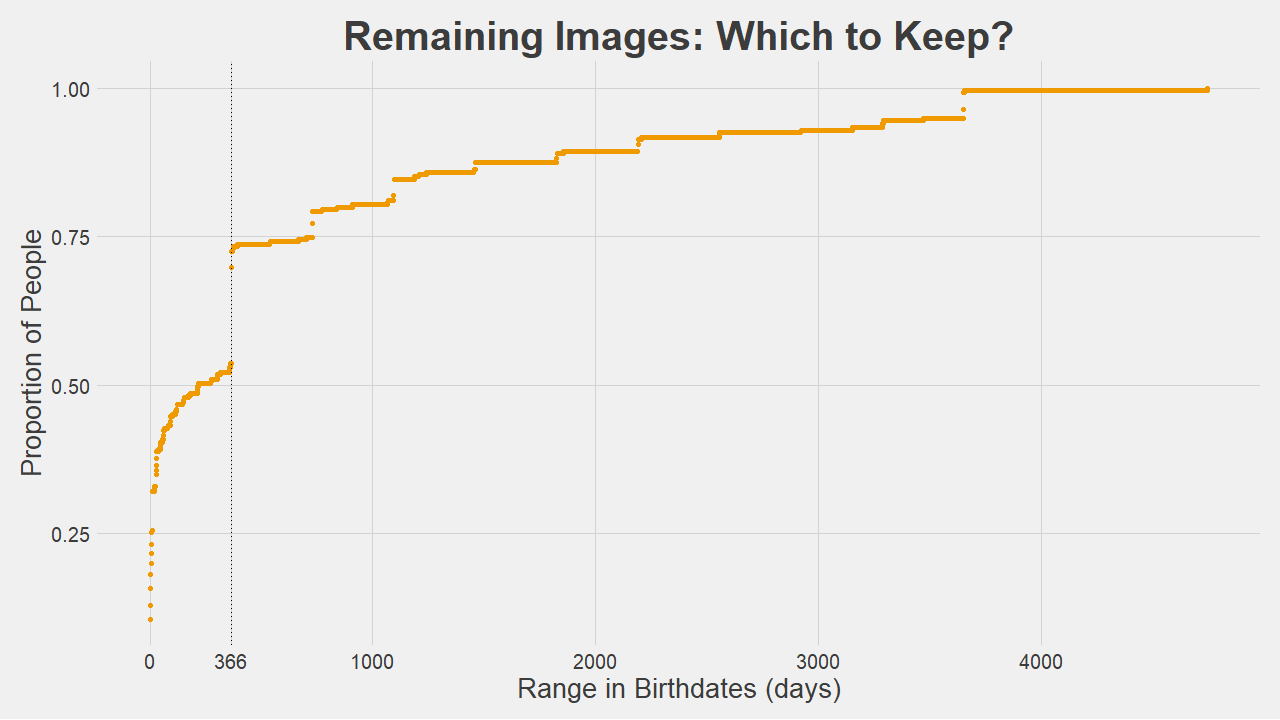}
\end{figure}

Figure \ref{age_inconsistent} displays the proportion of subjects' birthdates that are within a given range in days. For each subject whose birthdates differ by no more than one year (366 days to include leap year), we calculated the mean birthdate and assigned this date to all images of the subject. This strategy was applied to 185 subjects. Previous studies generally have used the floor age instead of the decimal age. For example, a subject who was actually 20.6 years old would be recorded as simply 20 years old. Thus, it is reasonable to use the mean birthdate when the range is within one year. The remaining 70 subjects' birthdates differ by more than one year. Because of the high number of such cases and the difficulty of human age perception, the associated 230 images were labeled as {\it dob - uncorrectable}. 

\subsection{Multiple Versions of Cleaned Datasets}

After cleaning MORPH-II of gender, race, and birthdate inconsistencies, three cleaned datasets were created:

\begin{itemize}
\item {\it morphII\_cleaned\_v2} - same as original dataset (morph\_2008\_nonCommercial.csv), but with all race, gender, and dob (except for images with \textit{dob - uncorrectable}) inconsistencies corrected 
\item {\it morphII\_go\_for\_age} - images with \textit{dob - uncorrectable} are removed from the above dataset. This leaves all the images with consistent age information that are ready for training and testing age estimation models
\item {\it morphII\_holdout\_for\_age} - contains only the images with \textit{dob - uncorrectable}.
\end{itemize}

Two new variables were created for each of the above datasets, including a new indicator variable (0-8) and age\_dec as decimal age (the difference between date of birth and date of arrest). 
The indicator variable takes a value between 0 and 8 representing the changes made to a given entry. The coding scheme for the indicator variable is described in Table \ref{table7}. This variable offers researchers the flexibility to easily include or omit data with certain types of inconsistencies.
There are N=55,134 data entries in the dataset of {\it morphII\_cleaned\_v2}, while N=54,904 in {\it morphII\_go\_for\_age} and N=230 in {\it morphII\_holdout\_for\_age}.


\begin{table}[ht]
\centering
\caption{Indicators for new variable {\it corrected} }
\label{table7}
\begin{adjustbox}{width=0.48\textwidth}
\begin{tabular}{c  l  l}
\toprule
\textbf{Indicator} & \textbf{Information}   & \textbf{\# of images}\\ \midrule
\textbf{0} & no change                      & {\bf 52,414}        \\
\textbf{1} & dob - majority                 & 1,906                \\
\textbf{2} & dob - averaged                 & 515                 \\
\textbf{3} & dob - uncorrectable            & 230                 \\
\textbf{4} & race - majority                &11                 \\
\textbf{5} & race - perception              &22                 \\
\textbf{6} & race - too difficult to tell, assigned to Other &33 \\
\textbf{7} & more than 1 change             & 2                 \\
\textbf{8} & gender corrected               &1 \\ \midrule
 & & Total = 55,134\\
\bottomrule
\end{tabular}
\end{adjustbox}
\end{table}

\subsection{Research Based on Uncleaned MORPH-II Data}


A substantial amount of research has been conducted using the MORPH-II dataset. Unfortunately, researchers reporting, for example, that the total number of subjects in the dataset is 13,618 (when it is actually 13,617),
is an indication that the data were not properly cleaned. Without discrediting the important contributions that have been made, such research outcomes could be more accurate if the data were cleaned properly.
%
%

\begin{figure}[h]
\centering
\caption{Frequency of subjects with inconsistent birthdates by number of days.} \bigskip
\label{age_inconsistent summary}
\includegraphics[width = 0.48\textwidth,keepaspectratio]{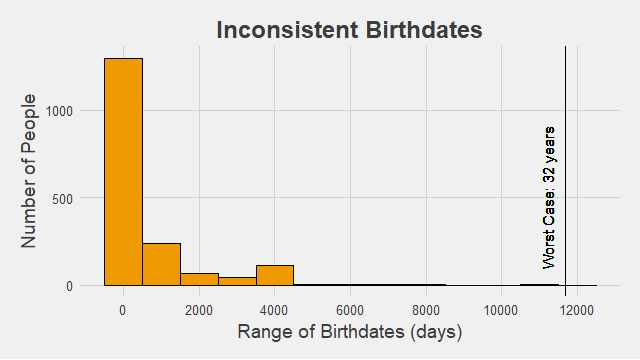}
\end{figure}
There will not likely be an enormous impact on model performance for gender or race classification, because the number of gender and race inconsistencies is relatively small. However, age estimation models may see a change in error. 
Figure \ref{age_inconsistent summary} summarizes the inconsistent birthdates. The worst case is a subject whose reported birthdates are 32 years apart, as displayed in Figure \ref{Worst Inconsistent Birthdates}. In some cases, subjects' birthdates change such that their reported age in the dataset decreases with time. The correction of such issues could significantly affect models concerned with age estimation or age progression.



\section{Automatic Subsetting Scheme }

As motivated in Section I, evaluating the performance of models through cross-validation can be challenging on MORPH-II. Guo and Mu \cite{ guo2010human, guo2010study, guo2011simultaneous} proposed a subsetting scheme as an evaluation protocol, which many subsequent studies followed. Their methods aim to overcome the disproportionate distributions of racial and gender groups by utilizing only images of black and white individuals. To maximize the number of represented females but without altering the original distributions too severely, they built two non-overlapping subsets $S1$ and $S2$ that each have a 3:1 male to female ratio and a 1:1 black to white ratio. Each subset is divided into gender groups first, which are further split by race, while maintaining similar age distribution. While $S1$ and $S2$ have non-overlapping images, the same subject (if arrested multiple times) might be pictured in both subsets. As a result, we seek to reconstruct subsets similar to $S1$ and $S2$, while guaranteeing that the subsets are truly independent (i.e., a subject is pictured in $S1$ or $S2$, but not both). 

It can be laborious to manually split the full data into folds with all the criteria outlined above. Thus, we explore ways to automatically split MORPH-II into independent folds according to the aforementioned evaluation protocol. This study can help inform future subsetting studies.

\subsection{Proposed Automatic Subsetting Scheme}
\subsubsection{Development}
Following Guo's and Mu's subsetting scheme, we only consider the white and black races for the training sets, since the number of images for other racial groups is too small. We also retain the gender and race ratios from their scheme. To automate subsetting, we use randomization to create many candidate subsets from which to choose. These possible subsets can be used for comparative purposes in the future; models could be built and validated on different subsets, and the results could be averaged or compared.


We use the cleaned version of MORPH-II with unidentifiable birthdates removed, referred to as {\it morphII\_go\_for\_age} in the previous section. Recall that this version includes the new variable {\it age dec}, which represents the exact age in decimal of the subject pictured in each image. This variable will be used in subsetting and future facial demographic studies hereafter, since the values are less biased than integer-valued ages. The decimal age values are also advantageous due to their improved continuity, which is essential as an assumption for the nonparametric tests considered in this paper.
%
For ease of comparison, we attempt to be as consistent as possible with the set notation in \cite{ guo2010human, guo2010study, guo2011simultaneous}. Let $W$ be the whole cleaned dataset ({\it morphII\_go\_for\_age}), $S$ the main training/validation set, and $R$ the remaining set. We divide $S$ into $S1$ and $S2$, such that $S1$ and $S2$ have the same number of images. We fix the ratios of male to female images as 3:1 and white to black images as 1:1.


Because white females are the smallest race-gender combination, we include all 2,570 white females in $S$. We randomly allocate each white female subject to either $S1$ or $S2$ exclusively, while constraining the total number of white female images in $S1$ to equal the number of white female images in $S2$.
Note that all white females are included in $S$, hence none are included in $R$.

For the other race-gender categories (black females, white males, and black males), we include only a portion of their images in $S$, while the remainder go in $R$. For black females, we randomly allocate a subset of subjects to $S1$ and an exclusive subset of subjects to $S2$, such that the total number of black female images in $S1$ is equal to the total number of black female images in $S2$, as well as equal to the total number of white female images in $S_i$ (for $i=1,2$). The images pertaining to any remaining black female subjects are sent to $R$.
%
For white males, we randomly allocate some subjects to $S1$ and other distinct white male subjects to $S2$, such that 3 times the number of white female images are in $S1$. The number of white male images in $S1$ is also set to be equal to the number of white male images in $S2$. Any remaining white males' images are sent to $R$.
%
The same process is repeated for black males, so that there are equal numbers of black male images within $S1$ and $S2$. The number of black male images is equal to the number of white male images, and other equalities hold too.

In this way, we ensure independence between $S1$, $S2$, and $R$. There is no expected information leakage between the training and testing sets. However, it should be clarified that observations within each set are not independent. For each subject in some set $\Omega$, all of the subject's images are in the same set $\Omega$. Hence, some observations within each set $\Omega$ are correlated with each other. 

\subsubsection{Implementation}
We implement our subsetting scheme in the statistical software R. We iterate through various random seeds $k= 1, 2, \hdots, l$ for some $l \in \mathbb{N}^+$. Subsets are randomly generated for each value of $k$. In this way, numerous candidate subsets are created. 
Among the candidate subsets, we seek those with similar age distributions. We obtain the age distributions of images in $S1$ and $S2$. Then for each value of $k$, we perform both the Anderson-Darling (AD) and Kolmogorov-Smirnov (KS) tests on those distributions. The hypotheses for both tests are as follows:
\begin{equation*}
\begin{aligned}
H_{o}&: S1_{age} \text{ has the same distribution as } S2_{age}
\\
H_{a}&: S1_{age} \text{ does not have the same distribution as } S2_{age}.
\end{aligned}
\end{equation*}

Both the AD \cite{darling1957kolmogorov, pettitt1976two} and the KS \cite{kolmogorov1933sulla} test are based on the empirical cumulative distribution function (ECDF) of data to examine whether two samples originate from two identical distributions. For a sample of $n$ observations $z_1$, $z_2$, $\hdots$ , $z_n$, the ECDF $\hat F_{n}(x)$ can be calculated as follows:
\begin{equation*}
\hat F_{n}(x) =
\begin{cases}
0, &\text{if $x < z_{(1)}$,}\\
i/n, &\text{if $ z_{(i)} \leq x < z_{(i+1)}$}\\
1, &\text{if $x \ge z_{(n)}$,}
\end{cases},
\end{equation*}
where $z_{(1)} < z_{(2)} <\hdots < z_{(n)}$ are the ordered sample. The ECDF $\hat G_{m}(x)$ for a sample of $m$ observations can be defined similarly. Let $N=n+m$ denote the total number of observations.

The two-sample AD test statistic is defined by:
\begin{equation*}
D_{AD}^{2}=\frac{nm} {N} \int_{-\infty}^{\infty} \frac{[\hat F_n(x) - \hat G_m(x)]^2} { H_N(x) [ 1-H_N(x)]} dH_N(x),
\end{equation*}
where $H_N(x) = (n \hat F_n(x) +m \hat G_m(x))/N $ is the ECDF of the pooled sample. If $H_N(x)=1$, then the integrand is defined as 0 conventionally.

The KS test statistic is the supremum (if attained) of the absolute values of the difference between the two ECDFs $\hat F_{n}(x)$ and $\hat G_{m}(x)$:
\begin{equation*}
D_{KS}=\sqrt{nm \over {N}} \sup\limits_{x} \{ \hat F_{n}(x) - \hat G_{m}(x)\}.
\end{equation*}
The KS test can be conducted as an exact permutation test. 

\begin{table}[h]
\centering
\caption{Number of Images in Subsets by Race and Gender}
    \label{table:images}
\begin{adjustbox}{width=0.5\textwidth}
	\begin{tabular}{l|cccccc|c|cc}
\hline
& \textbf{WF} & \textbf{BF}  & \textbf{WM} & \textbf{BM} & \textbf{dF} & \textbf{dM} &\textbf{Overall} &\textbf{F} & \textbf{M}  \\
\hline
\textbf{S1} & 1,285  & 1,285  & 3855 & 3,855 & 0 &0 & 10,280 & 2570    & 7,710\\
\textbf{S2} & 1,285  & 1,285   & 3,855     & 3,855 &0 & 0 &10,280 & 2,570     & 7,710 \\
\textbf{R}  & 0 & 3,150  & 220    & 28,980 &144 &1,850 & 34,344   & 3,294    &31,050 \\
\hline
\textbf{Overall}  & 2,570 & 5,720  & 7,930 & 36,690  &144 &1,850 & 54,904 & 8,434 & 46,470\\
\hline
\end{tabular}
\end{adjustbox}
\bigskip

    \caption{Number of Distinct Subjects in Subsets by Race and Gender}
    \label{table:subjects}
\begin{adjustbox}{width=0.5\textwidth}
	\begin{tabular}{l|cccccc|c|cc}
\hline
        & \textbf{WF} & \textbf{BF}  & \textbf{WM} & \textbf{BM} & \textbf{dF} & \textbf{dM}  &\textbf{Overall} & \textbf{F} & \textbf{M} \\
\hline
\textbf{S1} & 311  & 332  & 1,005    & 948  &0 &0 &2,596 & 643    & 1,953 \\
\textbf{S2} & 313  & 336   & 988     & 943 &0 &0  &2,580 & 649     & 1,931 \\
\textbf{R}  & 0 & 809  & 55    & 6,899  &40 &568 &8,371 &849  &7,522 \\
\hline
\textbf{Overall}  & 624 & 1,477  & 2,048 & 8,790 &40  &568  &13,547 &2,141  &11,406\\
\hline
\end{tabular}
\end{adjustbox}

\bigskip

\caption{Additional Race Groups in Remaining Subset R}
\label{table:race}
\begin{adjustbox}{width=0.5\textwidth}
\begin{tabular}{l|cccccc|c|cc}
\hline
& \textbf{HF} & \textbf{AF}  & \textbf{OF} & \textbf{HM} & \textbf{AM} & \textbf{OM} &\textbf{Overall}  &\textbf{F} &\textbf{M}\\
\hline
\textbf{Subjects in R} &28 &4 &8 &502 &47 &19 &608 &40 &568\\
\textbf{Images in R}  &99 &13 &32 &1,646 &140 &64 &1,994 &144 &1,850 \\
\hline
\end{tabular}
\end{adjustbox}

\bigskip

\caption{Numerical Summary of Age in Sets}
\label{table:agenum}
\begin{adjustbox}{width=0.5\textwidth}
\begin{tabular}{l|ccccc|cc}
\hline
& \textbf{Min.} & \textbf{Q1} & \textbf{Median} & \textbf{Q3} & \textbf{Max} & \textbf{Mean} & \textbf{SD}\\
\hline
\textbf{S1} & 16.003 & 24.296 & 34.495 & 42.185 & 77.196 & 34.041 & 10.957 \\
\textbf{S2} & 16.005 & 24.370 & 34.371 & 42.014 & 75.421 & 33.926 & 10.908\\
\textbf{W} & 16 & 23.369 & 33.091 & 41.422 & 77.196 & 33.019 & 10.950\\
\hline
\end{tabular}
\end{adjustbox}
\end{table}

\begin{figure}
\centering
\includegraphics[width=.5\textwidth]{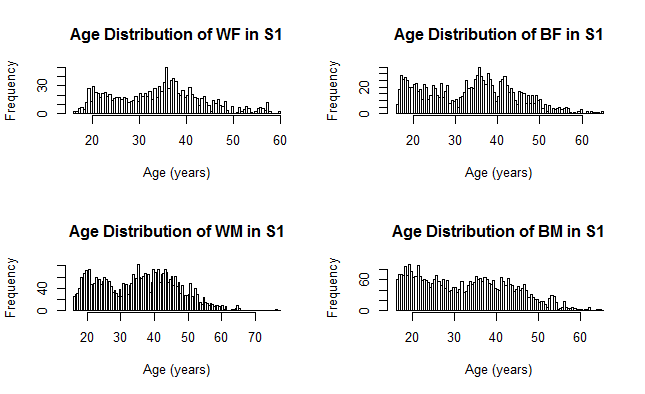}
\caption{Observed age histograms in $S1$ for random seed 42. }
\label{fig:agehistS1}
\bigskip
\includegraphics[width=.5\textwidth]{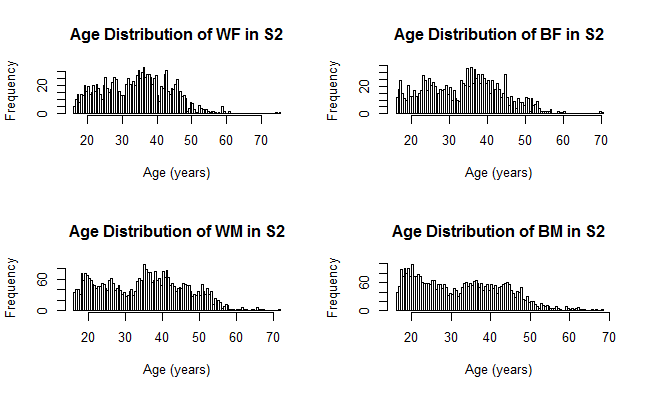}
\caption{Observed age histograms in $S2$ for random seed 42. }
\label{fig:agehistS2}

\bigskip

\includegraphics[width=0.45\textwidth]{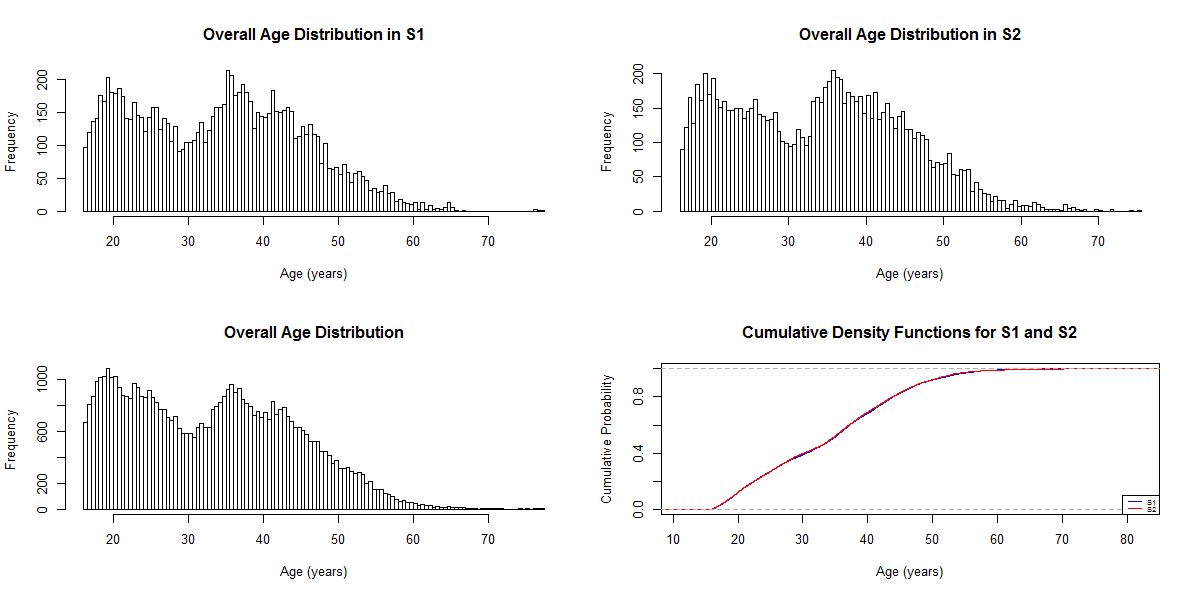}
\caption{Age distributions in $S1$, $S2$, and $W$ for random seed 42. }
\label{fig:ageall}
\end{figure}

We use the \textit{p}-values of both tests to identify the best subsets. High \textit{p}-values indicate $S1$ and $S2$ have similar age distributions for a particular seed $k$, so the \textit{p}-values can serve as metrics for judging suitable subsets. In this context, the \textit{p}-values are not to be interpreted as clear probabilities for the following reasons: not all assumptions for the tests are met (since some observations within each set are dependent) and the significance level cannot be defined appropriately when an indefinite number of tests are made. We believe our unconventional use of \textit{p}-values is valid here, since we are not attempting to make any probability statements based on them.

Using these criteria, we identify random seed number $k=42$ as one which produces satisfactory subsets. The \textit{p}-values for the KS and AD tests are $0.657$ and $0.652$, respectively. The statistical summaries below further indicate the suitability of these subsets. We do not guarantee that random seed $42$ produces the global optimum results, but it is found to be satisfactory for our purposes.

Tables \ref{table:images}, \ref{table:subjects}, and \ref{table:race} include basic information pertaining to the subsets generated by the random seed $42$.  
Table \ref{table:images} shows the number of images in subsets by race and gender, while Table \ref{table:subjects} gives the number of distinct subjects in subsets by race and gender. More detailed information on the different races is summarized in Table \ref{table:race}, with the number of images and the number of distinct subjects for the additional race groups in the remaining subset R. In Tables \ref{table:images} and \ref{table:subjects}, \textbf{d} denotes different race subjects (\textbf{H}ispanic, \textbf{A}sian, or \textbf{O}ther).

Additional graphical and numerical summaries are presented in Figures \ref{fig:agehistS1}, \ref{fig:agehistS2}, \ref{fig:ageall}, and Table \ref{table:agenum}.
Figure \ref{fig:agehistS1} displays the observed age histograms in $S1$. All the gender-race combinations in $S1$ have similar, right-skewed age distributions. Any differences in distribution seem minor and unlikely to significantly affect gender or race classification in future experiments. Figure \ref{fig:agehistS2} presents the observed age histograms in $S2$. Based on the plots, all the gender-race combinations in $S2$ are similarly distributed. Further, we see that the age distributions in $S1$ are not much different than the age distributions in $S2$. We do not expect any deviations in age distribution between sets to impact classification in a significant way.

The age distributions for images in the $S1$, $S2$, and $W$ subsets are depicted in Figure \ref{fig:ageall}. All three histograms are right-skewed with a roughly bimodal structure, indicating that $S1$ and $S2$ have been chosen successfully; the age distributions of images in the subsets $S1$ and $S2$ are close to the overall age distribution of images in $W$. The final plot shows the ECDFs of $S1$ and $S2$. It is difficult to distinguish the densities corresponding to each subset, since departures are so minor. This aligns with our expectations, because the $p$-values for the KS and AD tests were quite large (at approximately 0.65 for each).
In Table \ref{table:agenum}, the 5-Number Summary, as well as mean and standard deviation, are given for age in $S1$, $S2$, and $W$. The statistical summaries are nearly identical, further confirming these subsets' balanced age distributions.

All numerical and graphical summaries considered here indicate the suitability of the subsets generated from random seed $42$. These subsets are expected to yield good results for a variety of face imaging tasks, including gender and race classification, as well as age regression. 

\section{Posterior Probabilities for Race-Composite Age Estimation}

Previous research with MORPH-II considers distinct age estimation models for race and gender subgroups. However, these models produce estimates as if subjects belong to only one race category. For example, a subject might be predicted as Black with probability $0.6$ and as White with probability $0.4$. Then the subject might be classified as Black, and the age of the individual would be predicted using a race-dependent model. However, here we exploit the probabilistic racial information in the hope of producing more accurate estimates of age; e.g., two estimates of age would be computed conditional on race (White or Black) and weighed by $0.6$ and $0.4$, respectively. The proposed framework is displayed in Figure \ref{overview} and discussed below. 
\input{Flowchar_Age1.tex}



\subsection{Pre-Processing, Feature Extraction, and Dimension Reduction}
The face images were preprocessed as follows. First, the images were converted to grayscale. The faces were detected and aligned then cropped to $60 \times 70$ pixels, ensuring that the eyes were approximately 15 pixels from the side of the image and 25 pixels from the top of the image. For feature extraction, local binary patterns (LBPs) \cite{he1990texture} were calculated. In summary, this process involves dividing the image into fixed-size grids of pixels. Within each grid, the surrounding pixels are compared to the center pixel and assigned a value of zero or one, depending on whether they are lighter or darker; the corresponding binary string is converted to a base-10 integer, and a histogram of the values is created. Finally, the histograms (generated from each grid) are concatenated to form a feature vector for the image.
Next, the feature vectors are transformed and reduced to dimension $400$ through principal component analysis (PCA), one of the most popular dimension reduction methods in statistics. 
Briefly, PCA finds a linear transformation of the data such that maximal variation is contained in the initial transformed variables \cite{jolliffe2011principal}.


\subsection{Combining Race Posterior Probabilities with Age Estimates}
For the race classifier, a support vector machine (SVM) with a linear kernel was chosen \cite{cortes1995support}. Support vector regression (SVR) with a linear kernel was selected for age prediction \cite{drucker1997support}. 
The main parameter required by SVM and SVR is a cost parameter that controls the extent to which misclassifications of observations are permitted. For each model, we tuned this parameter through a multi-tiered grid search.

For each subject in a fixed gender class, the probability of race (White or Black) was calculated from a modification developed by \cite{platt} to the SVM output. For example, the posterior probability a subject is white was calculated as follows: 

\begin{equation}
P( \textnormal{Race}=White \mid f )= \frac{1}{1 + \exp(Af + B)}, 
\end{equation}
where $f\in \mathbb{R}$ is the SVM output, and $A$ and $B$ are coefficients determined by fitting (1) to the training data \cite{platt}. For each image, let $w$ and $b$ denote the posterior probabilities of being white and black, respectively. 

Still within each gender class, two independent models to predict age were fit: SVR on black images and SVR on white images. For each image within the gender class, two predictions were obtained: the predicted age of the subject trained on white images and the predicted age of the subject trained on black images. Denote each estimate as $\hat{Y}_W$ and $\hat{Y}_B$, respectively. 

We define the race-composite age estimate $Y^\ast$ as the weighted mean of the separate age estimates for race:

\begin{equation}
Y^\ast = (b \cdot \hat{Y}_B) + (w \cdot \hat{Y}_W).
\end{equation}
This is the final age estimate of the pictured individual.


\input{Flowchar_Age2.tex}

\subsection{Experimental Design}
We discuss adapting the above methods to cross-validation on MORPH-II.
As noted in Section III, $S1$ and $S2$ are independent of one another and contain no common individuals. Hence, training of the race classifier and age estimator were done on the appropriate subsets of one main set, while testing was performed on the relevant subsets from the other main set. For example, when testing the model on bf\_1, a subset of black females, the race classifier was trained on f2, and the age models were trained on bf\_2 and wf\_2. The subsetting scheme is shown in Figure \ref{subset}.

\subsubsection{Toy Example}
A Toy Example was used to assess the model before applying it to a larger subset of MORPH-II; it contains 1,000 images of 1,000 distinct individuals (i.e., 1 image per subject) and is divided according to the subset scheme in Figure \ref{subset}. 
It differs from the full set MORPH-II in that the overall age distribution is uniform. The results obtained on this dataset, displayed below in Table \ref{Age_estimation}, show an improvement in mean absolute error (MAE) with the use of posterior probability. Note that weighted MAE is compared to the MAE when age is predicted using only the age model for the race with the highest associated posterior probability.
\begin{table}[ht]
\centering
\caption{Experiment Results on MORPH-II: 1=Toy; 2=$S$}
\begin{tabular}{c|cc|cc}
\toprule
               & \textsc{MAE$^1$} & \textsc{Weighted MAE$^1$}  & \textsc{MAE$^2$} & \textsc{Weighted MAE$^2$}\\\midrule
bf\_1 & 6.71         & \textbf{6.695}    &6.894     & \textbf{6.872}                  \\
bf\_2 & 6.634        & \textbf{6.45}     & \textbf{6.316} & 6.389                      \\
wf\_1 & 7.007        & \textbf{6.535}    &5.826          & \textbf{5.824}              \\
wf\_2 & 6.579        & \textbf{6.461}    & \textbf{5.791} & 5.863                      \\
bm\_1 & 5.501        & \textbf{5.475}    &  5.283          & \textbf{5.279}           \\ 
bm\_2 & 4.979        & \textbf{4.799}    &  \textbf{5.359} & 5.362                     \\
wm\_1 & 4.443        & \textbf{4.434}    &  4.985          & \textbf{4.983}             \\
wm\_2 & 4.518        & \textbf{4.498}	& \textbf{4.971} & 4.987	\\\bottomrule       
\end{tabular}
\label{Age_estimation}
\end{table}

\subsubsection{MORPH-II Subset $S$}
The same experiment as before was performed on $S$, a set of 20,560 MORPH-II images containing black and white individuals (previously discussed in Section III). 
However, these results were not as conclusive as those from the partial dataset. As can be seen in Table \ref{Age_estimation}, MAE did not consistently improve with the use of the posterior probabilities of race.

\subsection{Results}

While the results from the partial dataset indicate the race-composite age estimate is an improvement over the single-race prediction, the preliminary results from the MORPH-II subset $S$ are mixed. In part, this may be caused by the different age distribution between the datasets; while the partial dataset has a uniform age distribution, $S1$ and $S2$ in $S$ have distributions that mimic the overall age distribution of MORPH-II. Additionally, because the race labels reflect only one race categorization, multiracial individuals in MORPH-II are labeled as only black or only white. Because of this, the training sets for all the models likely include mixed race individuals, presenting an issue for both race classification and age estimation. For example, an individual in the training set labeled as Black may actually be 70\% black and 30\% white. When the race classifier is trained on this individual, it may learn to classify all mixed race people as Black. Then the posterior probability produced from this classifier may not reflect the 70:30 racial composition of this individual, leading to an improper weighting of the race-composite age estimate. Further work will be required to investigate these issues, but race-composite age prediction remains hopeful for reducing error rates and accommodating mixed-race individuals. 

\section{Conclusion}
In our first preliminary study, we perform data validation and cleaning on the aging database MORPH-II. This is a necessary step to identify and correct inconsistencies, including contradictory race, gender, and age information for subjects with multiple entries in the dataset. The correction of such inconsistencies is critical for demographic analysis, including gender classification, race classification, and age prediction. To our best knowledge, no prior studies on MORPH-II have involved such extensive data validation and cleaning. Creating multiple versions of the cleaned data, we have improved the accuracy of this popular dataset and enabled researchers to save time in the future.  

In our second preliminary study, we propose an automatic subsetting scheme of MORPH-II. Our scheme is inspired by the work of Guo and Mu. \cite{ guo2010human, guo2010study, guo2011simultaneous}, but we do make some changes. Most notably, we maintain their racial and gender proportions, while ensuring independence between training and testing sets. Our approach is also novel in its random generation of various candidate subsets, which are selected based on the Kolmogorov-Smirnov and Anderson-Darling goodness-of-fit tests. We present a suitable choice of subsets for a specific random seed, but the generation of other subsets from random seeds are recommended for comparative purposes in the future. For any models built and tested using the subsetting scheme proposed in this study, we expect the estimates of test error or accuracy to be less biased than previous subsetting schemes. Our automatic subsetting scheme can be used for face imaging tasks involving gender, race, and age. Even though the automatic subsetting scheme is illustrated on the MORPH-II dataset, it can be extended to other experimental designs.

Our third preliminary study proposes a novel race-composite framework for age prediction. The results on a small subset of MORPH-II are promising, motivating continued study of this framework. It is expected to be most successful on images of individuals with correct race labels (e.g., full race information for mixed-race individuals). Improvements may also be possible with exploration of other feature extraction, dimension reduction, classification, and regression techniques. Beyond MORPH-II, the framework can be applied to other multiracial face image datasets.



\section{Acknowledgements}
This research was done through the National Science Foundation under DMS Grant Number 1659288. A special thanks to Morgan Ferguson, Catherine Nansalo, Kevin Park, and Rachel Towner for their support.

%
%

{\small
\bibliographystyle{IEEEtran}
\bibliography{References.bib}

\end{document}

%% file: Inconsistent_image.tex
\begin{figure*}[h]
\centering
\begin{minipage}{0.33\textwidth}
\centering
\caption{Gender Inconsistency} \bigskip
\label{gender_inconsistent}
\begin{tabular}{ccc}
\includegraphics[width = 0.55in,keepaspectratio]{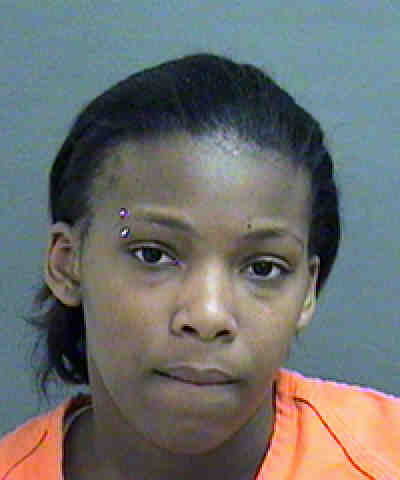} &
\includegraphics[width = 0.55in,keepaspectratio]{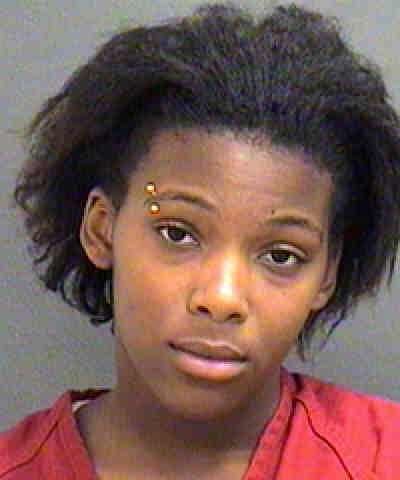} &
\includegraphics[width = 0.55in,keepaspectratio]{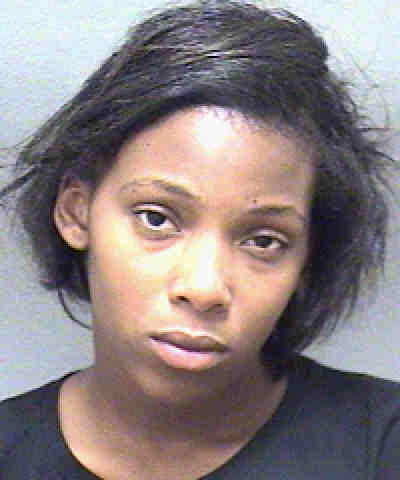} \\
\vspace{0.5cm}
{\footnotesize (a) Female} & {\footnotesize (b) Male} & {\footnotesize (c) Female} \\
\includegraphics[width = 0.55in,keepaspectratio]{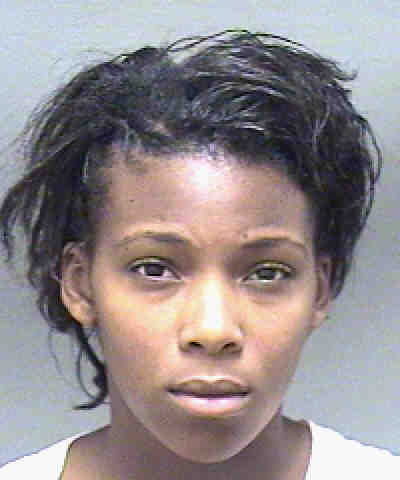} &
\includegraphics[width = 0.55in,keepaspectratio]{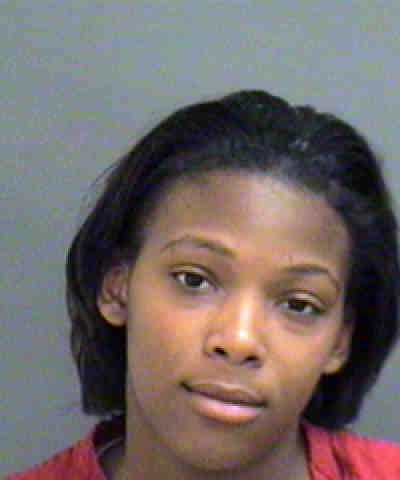} &
\includegraphics[width = 0.55in,keepaspectratio]{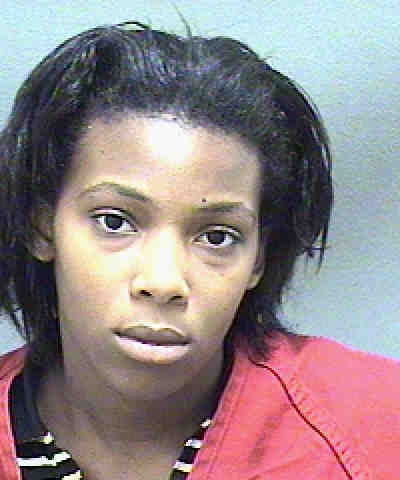} \\
{\footnotesize (d) Female} & {\footnotesize (e) Female} & {\footnotesize (f) Female}
\end{tabular}
\end{minipage}%
\vrule{}
\begin{minipage}{0.33\textwidth}
\centering
\caption{Race Inconsistencies} \bigskip
\label{race_inconsistent}
\begin{tabular}{ccc}
\includegraphics[width = 0.55in,keepaspectratio]{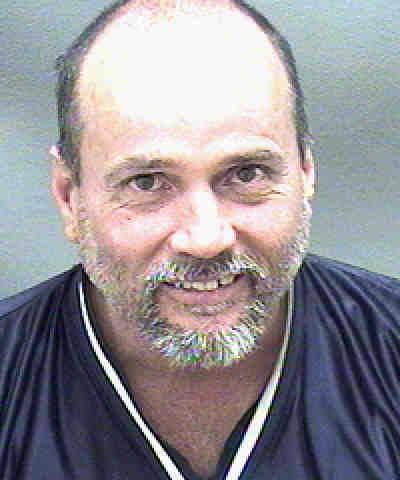} &
\includegraphics[width = 0.55in,keepaspectratio]{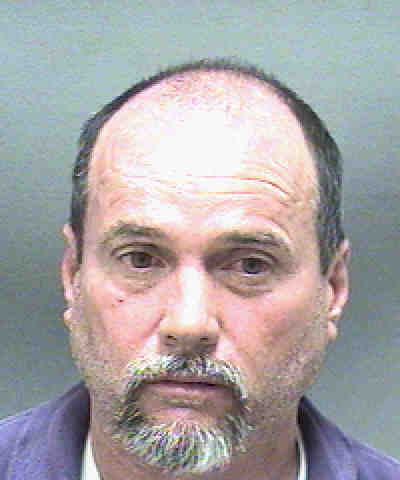} &
\includegraphics[width = 0.55in,keepaspectratio]{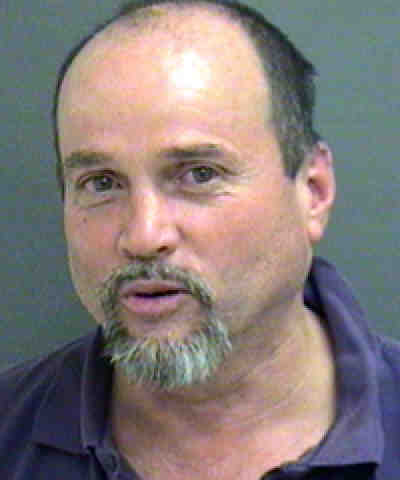} \\
\vspace{0.5cm}
{\footnotesize (1a) White} & {\footnotesize (1b) Black} & {\footnotesize (1c) White} \\
\includegraphics[width = 0.55in,keepaspectratio]{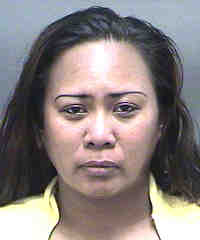} &
\includegraphics[width = 0.55in,keepaspectratio]{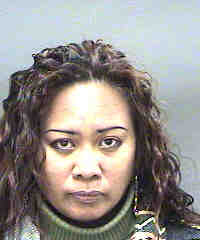} &
\includegraphics[width = 0.55in,keepaspectratio] {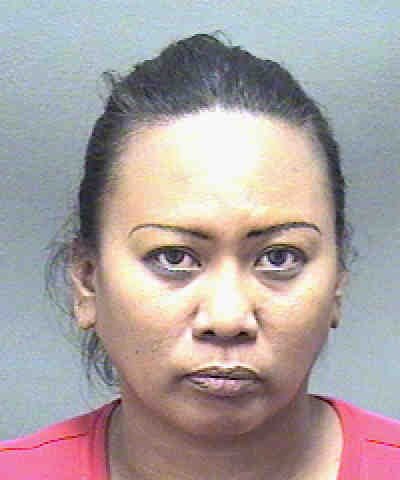} \\
{\footnotesize (2a) Asian} & {\footnotesize (2b) White} & {\footnotesize (2c) Black}
\end{tabular}
\end{minipage}
\vrule{}
\begin{minipage}{0.32\textwidth}
\centering
\centering
\caption{Worst Inconsistent Birthdates} \bigskip
\label{Worst Inconsistent Birthdates}
\begin{tabular}{ccc}
\includegraphics[width = 0.55in,keepaspectratio]{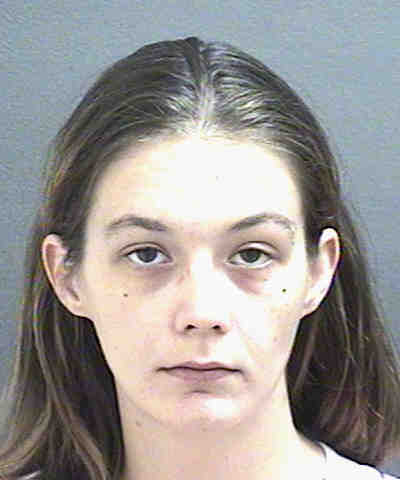} &
\includegraphics[width = 0.55in,keepaspectratio]{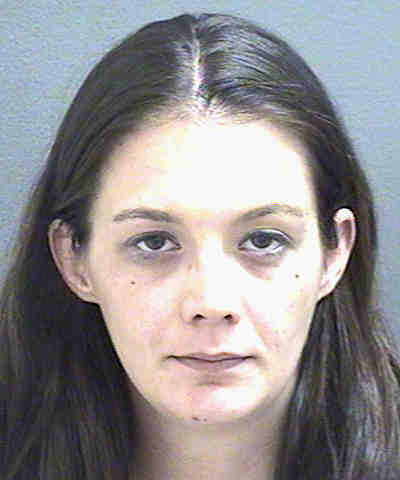} &
\includegraphics[width = 0.55in,keepaspectratio]{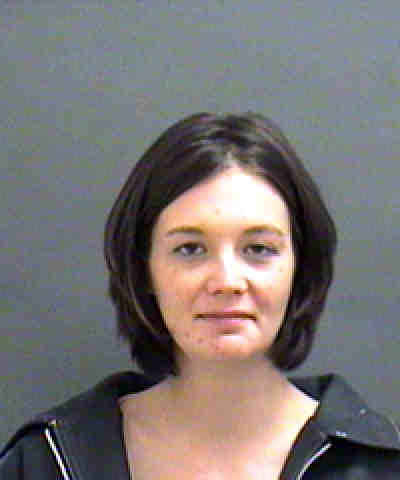} \\
\vspace{0.5cm}
{\footnotesize (1a) Age=23} & {\footnotesize (1b) Age=55} & {\footnotesize (1c) Age=23} \\
\includegraphics[width = 0.55in,keepaspectratio]{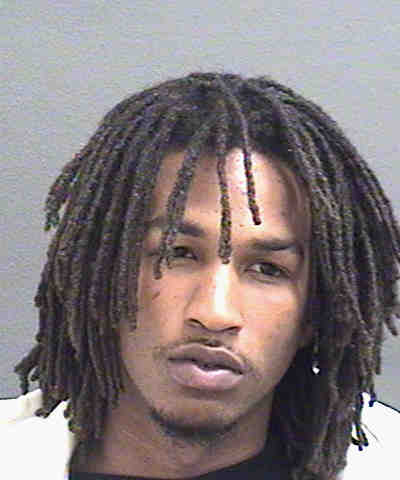} &
\includegraphics[width = 0.55in,keepaspectratio]{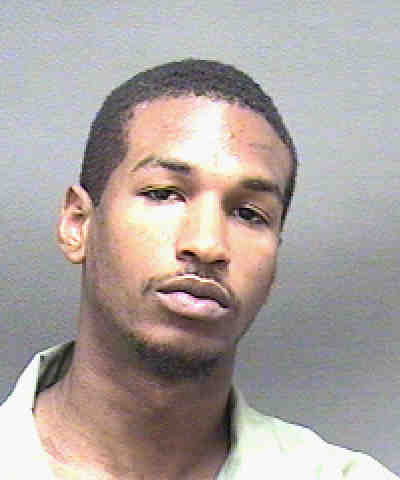} &
\includegraphics[width = 0.55in,keepaspectratio]{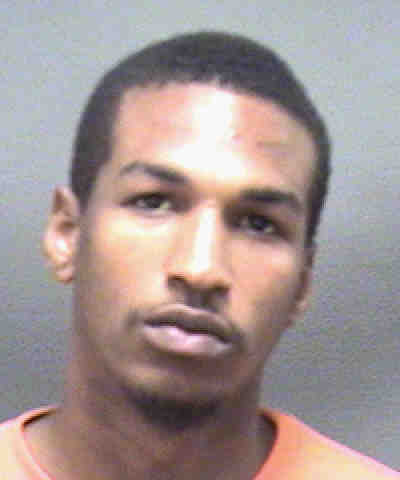} \\
{\footnotesize (2a) Age=20} & {\footnotesize (2b) Age=21} & {\footnotesize (2c) Age=51} \\
\end{tabular}
\end{minipage}
\end{figure*} 

%% file: Flowchar_Age1.tex
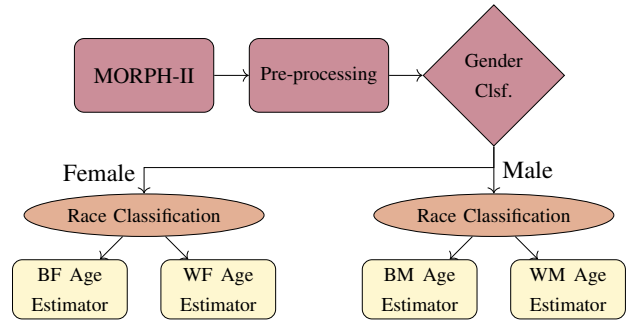
\begin{figure}[!ht]  
\centering
\begin{adjustbox}{max width=.45\textwidth}
  \begin{tikzpicture}[node distance = 3cm, auto]
  
    \node [block] (init) {\small MORPH-II};
    \node [block, right of = init, node distance = 2.5cm] (pp) {\footnotesize Pre-processing};
    \node [decision, right of = pp, node distance = 2.5cm] (gc) {\footnotesize Gender Clsf.};
    \node [cloud, below of = gc, node distance = 2cm] (m) {\footnotesize Race Classification};
    \node [cloud, below of = init, node distance = 2cm] (f) {\footnotesize Race Classification};
    
    \node [block3, below left of = m, node distance = 1.5cm] (of) {\footnotesize BM Age Estimator};
    \node [block3, below right of = m, node distance = 1.5cm] (hf) {\footnotesize WM Age Estimator};
    \node [block3, below left of = f, node distance = 1.5cm] (om) {\footnotesize BF Age Estimator};
    \node [block3, below right of = f, node distance = 1.5cm] (hm) {\footnotesize WF Age Estimator};
    \draw[->] (init) -- (pp);
    \draw[->] (pp) -- (gc);
    \draw[->] (gc) -- node {Male} (m);
    \draw[->] (gc.south) -- ++(0,-.3cm) -| (f) node[near end, above left, yshift=-4pt] {Female};
 
    \draw[->] (m) -- (of);
    \draw[->] (m) -- (hf);
    
    \draw[->] (f) -- (om);
    \draw[->] (f) -- (hm);
    
  \end{tikzpicture}
  \end{adjustbox}
\caption{Overview of race-composite age prediction framework. The four subgroup age estimators correspond to black females (BF), white females (WF), black males (BM) and white males (WM).}
\label{overview}
\end{figure}

%% file: Flowchar_Age2.tex
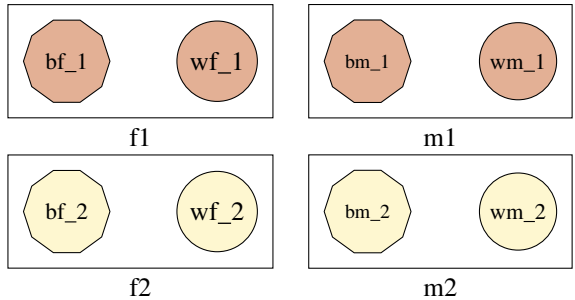
\begin{figure}[t]
  \centering
  \begin{adjustbox}{max width=.45\textwidth}
 \begin{tikzpicture}[node distance = 2cm, auto]
    \node [black,fill=BurntOrange!50, scale=.88] (bf1) {bf\_1};
    \node [white, right of=bf1, fill=BurntOrange!50] (wf1) {wf\_1};
    \node [black, right of=wf1, fill=BurntOrange!50, scale=.75, node distance=2cm] (bm1) {bm\_1};
    \node [white, right of=bm1, fill=BurntOrange!50, scale=.85] (wm1) {wm\_1};
    
    \node [black, below of=bf1, fill=Goldenrod!20, node distance=2cm, scale=.88] (bf2) {bf\_2};
    \node [white, below of=wf1, fill=Goldenrod!20, node distance=2cm] (wf2) {wf\_2};
    \node [black, below of=bm1, fill=Goldenrod!20, scale=.75, node distance=2cm] (bm2) {bm\_2};
    \node [white, below of=wm1, fill=Goldenrod!20, node distance=2cm, scale=.85] (wm2) {wm\_2};
    
    \node[draw,inner sep=2mm,label=below:f1,fit=(bf1) (wf1)] {};
    \node[draw,inner sep=2mm,label=below:f2,fit=(bf2) (wf2)] {};
    \node[draw,inner sep=2mm,label=below:m1,fit=(bm1) (wm1)] {};
    \node[draw,inner sep=2mm,label=below:m2,fit=(bm2) (wm2)] {};
    
  \end{tikzpicture}
  \end{adjustbox}
  \caption{Subsetting scheme for race-composite age prediction. Colors denote $S1$ and $S2$.}
  \label{subset}
\end{figure}